# A Comparative Study of Histogram Equalization Based Image Enhancement Techniques For Brightness Preservation And Contrast Enhancement

Omprakash Patel, Yogendra P. S. Maravi and Sanjeev Sharma

School of Information Technology, RGPV, Bhopal, Madhya Pradesh, India

## ABSTRACT

*Histogram Equalization is a contrast enhancement technique in the image processing which uses the histogram of image. However histogram equalization is not the best method for contrast enhancement because the mean brightness of the output image is significantly different from the input image. There are several extensions of histogram equalization has been proposed to overcome the brightness preservation challenge. Contrast enhancement using brightness preserving bi-histogram equalization (BBHE) and Dualistic sub image histogram equalization (DSIHE) which divides the image histogram into two parts based on the input mean and median respectively then equalizes each sub histogram independently. This paper provides review of different popular histogram equalization techniques and experimental study based on the absolute mean brightness error (AMBE), peak signal to noise ratio (PSNR), Structure similarity index (SSI) and Entropy.*

## 1. INTRODUCTION

Contrast enhancement is an important area in the field of digital image processing for human visual perception and computer vision. It is extensively used for medical image processing and as a preprocessing step in speech recognition, texture synthesis, and many other image/video processing applications [1] [2] [3]. There are many different methods have been developed for image contrast enhancement [4]-[22]. Here we discussed some popular image contrast enhancement method for brightness preservation.

Contrast Enhancement is a specific characteristics enhancement in the image enhancement processing. In which histogram equalization (HE) is the one of the popular method of image contrast enhancement [4]. The histogram of the discrete gray-level image represent the frequency of occurrence of all gray-levels in the image [5]. Histogram equalization well distributes the pixels intensity over the full intensity range. But HE has "mean-shift" problem [6], it shifts the mean intensity value to the middle gray level of the intensity range. So this technique is not useful where brightness preservation is required. To overcome this problem BBHE (Brightness Preserving by Histogram Equalization) has been proposed by Yeong-Taeg Kim [6] which overcome the mean shift problem. BBHE divides the input image histogram into two parts based on the mean value. Then histogram equalization is applied to the both separated parts with the new intensity range, first minimum gray level to the mean value and for second mean value to the maximum gray level.

Later, equal area dualistic sub-image histogram equalization (DSIHE) introduced by Yu Wan, Qian Chen and Bio-Min Zang [7] and claimed that the performance of DSIHE is better than BBHE in terms of brightness preservation and entropy. DSIHE uses the same concept as BBHE, it divides the input image histogram into two parts based on cumulative probability density equal to 0.5 instead of mean value as in BBHE. It divides the image in two parts with equal size which contains equal number of pixel.

The extension of BBHE is RMSHE which perform recursively separation of input image histogram based on mean value of input histogram to the satisfactory number of recursion levels. Each new sub-histogram is separated based on respective mean. As the number of recursion level increases the mean brightness of output image is closer to the input image mean brightness. The recursive nature of RMSHE implies scalable brightness preservation which is very useful in consumer electronic products [8]. A new method RSIHE is a novel extension of DSIHE and used some characteristics of RMSHE. The RSIHE recursively separated the input image histogram based on the cumulative distribution function (cdf) equal to 0.5. The RSIHE separates each new sub histogram recursively to the specified satisfactory recursion level [10].

Chen and Ramil proposes Minimum Mean Brightness Error Bi-Histogram Equalization (MMBEBHE), which is an extension of BBHE which offers maximum brightness preservation. MMBEBHE method proposes to separate histogram into two sub-parts using threshold level which is responsible for minimum absolute mean brightness error (AMBE) [9].

Mary Kim and Min Gyo Chung proposed another new method RSWHE (Recursively Separated and Weighted Histogram Equalization) for brightness preservation and contrast enhancement. The RSWHE offers better results in comparison to previous methods in all aspects. The RSWHE segments the input histogram into two or more sub histogram recursively based on the mean or median as used in RMSHE and RSIHE, then apply the weighting process based on normalized power low function to modify the each sub-histogram. And then perform histogram equalization to each weighted output sub-histogram independently [12].

## A. HISTOGRAM EQUALIZATION-

Histogram equalization is contrast enhancement technique in a spatial domain in image processing using histogram of image. Histogram equalization usually increases the global contrast of the processing image. This method is useful for the images which are bright or dark.

**Implementation**

Consider the discrete grayscale input Image $X = x(i,j)$, with the $L$ discrete levels, where $x(i,j)$ represents the intensity levels of the image at the spatial domain $(i,j)$ Let histogram of Image $X$ is $H(X)$. Now the probability density function $pdf(X)$ can be defined as-

$$pdf(X_k) \, or \, p(X_k) = {}^{n_k}/_{N} \tag{1}$$

Where, $0 \le k \le (L-1)$

$L$ is the total number of gray levels in the image,

$N$ is the Total number of pixels in the image,

$n_k$ is the total number of pixels with the same intensity level $k$.

From the $pdf(X)$ (1) the cumulative distribution function $cdf(X_i)$ is defined as-

$$cdf(X_i) \, or \, c(X_i) = \sum_{i=0}^{k} p(X_i) \tag{2}$$

Note that $cdf(X_{L-1}) = 1$ from eq. (1) and (2).

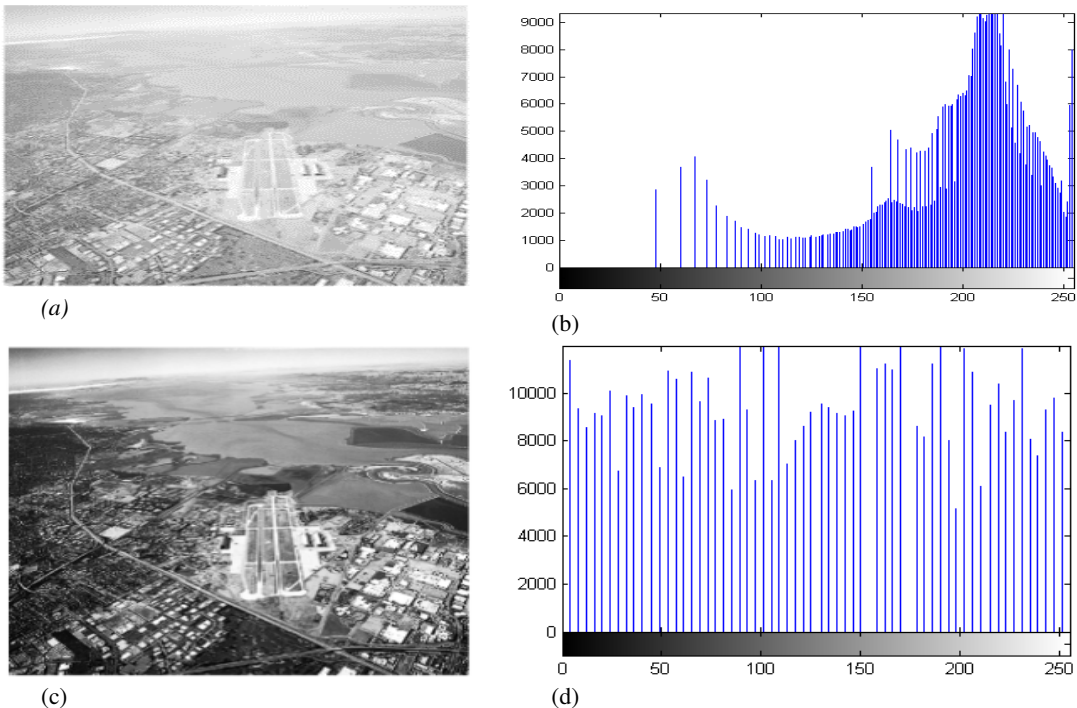

*(a)*

(b)

(c)

(d)

Fig. 1. (a) shows input image having low contrast, (b) shows histogram of input image, (c) shows histogram equalized image and (d) shows histogram of processed image

Histogram equalization is a scheme that maps the input image into the entire dynamic range $[X_0, X_{L-1}]$ by using the cumulative distribution function as a transform function. Let's define the transform function $f(X)$ using cumulative distribution function $cdf(X_i)$ as-

$$f(X) = X_0 + (X_{L-1} - X_0) \times cdf(X_i) \qquad (3)$$

Then the output image of histogram equalization, $Y = y(i,j)$ can be expressed as

$$Y = f(X) \qquad (4)$$

$$Y = \{f(x(i,j)) | \forall x(i,j) \in X\} \qquad (5)$$

The above describe the histogram equalization on gray scale image. However it can also be used on color image by applying the same method separately to the Red, Green and Blue Component of the RGB color image.

## B. Brightness Preserving Bi-Histogram Equalization (BBHE)-

The BBHE firstly decomposes an input image histogram $H(X)$ into two sub-images based on the mean of the input image. Now let input image $X$ is decomposed into two sub-images $X_L$ and $X_U$ by using the input mean $X_m$, where $X_m \in \{X_0, X_1, X_2 \ldots X_{L-1}\}$.

$$X = X_L \cup X_U \qquad (6)$$

where

$$X_L = \{x(i,j) | x(i,j) \leq X_m \ \forall \ x(i,j) \ \in X\}$$

and

$$X_U = \{x(i,j) | x(i,j) > X_m \ \forall \ x(i,j) \ \in X\}$$

And mean of the input image is calculated by,

$$X_m = \frac{\left(\sum_{i=0}^{L-1} i.p(X_i)\right)}{\left(\sum_{i=0}^{L-1} p(X_i)\right)} \qquad (7)$$

Note that the sub-image $X_L$ is composed of $\{X_0, X_1, X_2 \ldots X_m\}$ and sub-image $X_U$ is composed of $\{X_{m+1}, X_{m+2}, X_{m+3} \ldots X_{L-1}\}$

Next, define the respective probability density function (*pdf*) of the sub-image $X_L$ and $X_U$ as-

$$pdf_L(X_k) or \ p_L(X_k) = {n_k}/{N_L} \qquad (8)$$

$where, k = 0, 1, 2 \ldots m.$
and

$$pdf_U(X_k) \, or \, p_U(X_k) = {}^{n_k}/_{N_U} \qquad\qquad (9)$$
$$where, k = m+1, m+2, \dots L-1.$$

Where $N_L$ and $N_U$ represents the total number or pixels in the $X_L$ and $X_U$ respectively. The respective cumulative distribution function *(cdf)* is-

$$cdf_L(X_i) = \sum_{i=0}^{m} P_L(X_i) \qquad\qquad (10)$$

and

$$cdf_L(X_i) = \sum_{i=m+1}^{L-1} P_U(X_i) \qquad\qquad (11)$$

Where $X_i = x$. note that $cdf_L(X_m) = 1$ and $cdf_L(X_{L-1}) = 1$ by definition.

Now defining transformation function as defined in histogram equalization for each sub image-

$$f_L(X_L) = X_0 + (X_m - X_0)cdf_L(X_i) \qquad\qquad (12)$$

and

$$f_U(X_U) = X_{m+1} + (X_{L-1} - X_{m+1})cdf_U(X_i) \qquad\qquad (13)$$

Based on these transformation functions, sub images are equalized independently and the output of BBHE is composition of equalized sub images. Now the output of BBHE, *Y* can be expressed as-

$$Y = \{Y(i,j)\}$$
$$= f_L(X_L) \cup f_U(X_U) \qquad\qquad (14)$$

where

$$f_L(X_L) = f_L(X(i,j))|\forall X(i,j) \in X_L \qquad\qquad (15)$$

and

$$f_U(X_U) = f_U(X(i,j))|\forall X(i,j) \in X_U \qquad\qquad (16)$$

$f_L(X_L)$ Equalize the sub-image $X_L$ over the range $[X_0, X_m]$ whereas $f_U(X_U)$ equalize the subimage $X_U$ over the range $[X_{m+1}, X_{L-1}]$.

## C. Equal Area Dualistic Sub-Image Histogram Equalization (DSIHE):-

Equal Area Dualistic Sub-Image Histogram Equalization (DSIHE) follows the same idea as in the BBHE, which decomposes the original image histogram into two sub histogram based on the mean value. DSIHE method decomposes the image based on the gray level with a cumulative distribution value equals to 0.5 instead of the men as in BBHE method. DSIHE method decomposes the image aiming at the maximization of the Shannon's entropy of the output image. For such aim the input image is decomposed into two sub-images, being one dark and one bright, and then applies the HE on the decomposed sub-images. Then compose the resultant histogram equalized sub-image into one image to produce the DSIHE output image [7]. Assuming $X_D$ is the median value which divides input image into two parts $X_L$ and $X_U$ then $X_L$ and $X_U$ is be defined as-

$$X = X_L \cup X_U$$
$$(17)$$

where

$$X_L = \{x(i,j) \mid x(i,j) \leq X_D \; \forall \; x(i,j) \in X\}$$

and

$$X_U = \{x(i,j) \mid x(i,j) > X_D \; \forall \; x(i,j) \in X\}$$

and

$$X_D = \arg Min_{0 \leq k \leq L-1} \left| cdf(X_k) - \frac{cdf(X_0) + cdf(X_{L-1})}{2} \right| \tag{18}$$

Where sub-image $X_L$ is composed by the $\{X_0, X_1, \ldots, X_{e-1}\}$, and sub-image $X_U$ is composed by the $\{X_e, X_{e+1}, \ldots, X_{L-1}\}$. Now let's probability density function for sub-images $X_L$ and $X_U$ respectively are-

$$pdf_L(X_k) \, or \, p_L(X_k) = {n_k}/{N_L} \, , where \; k = 0,1,2 \ldots D. \tag{19}$$

and

$$pdf_U(X_k) \, or \, p_U(X_k) = {n_k}/{N_U} \, , where \; k = D+1, D+2, D+3 \ldots L-1. \tag{20}$$

Cumulative distribution function for both sub-images $X_L$ and $X_U$

$$cdf_L(X_i) = \sum_{i=0}^{D-1} P_L(X_i) \tag{21}$$

and

$$cdf_L(X_i) = \sum_{i=D}^{L-1} P_U(X_i) \tag{22}$$

Now the transform function is defined as-

$$f_L(X_L) = X_0 + (X_{D-1} - X_0)cdf_L(X_i) \tag{23}$$

and

$$f_U(X_U) = X_D + (X_{L-1} - X_D)cdf_U(X_i) \tag{24}$$

And output image

$$Y = \{Y(i,j)\}$$
$$= f_L(X_L) \cup f_U(X_U) \tag{25}$$

$f_L(X_L)$ Equalize the sub-image $X_L$ over the range $[X_0, X_{D-1}]$ whereas $f_U(X_U)$ equalize the subimage $X_U$ over the range $[X_D, X_{L-1}]$.

## D. Recursive Mean Separate Histogram Equalization (RMSHE):-

This method proposes the generalization of the Brightness preserving bi-histogram equalization (BBHE) to overcome such limitation and provide not only better but also scalable brightness preservation. BBHE separates the input histogram into two based on its mean before equalizing then independently while the separation is done only once in BBHE. The RMSHE method proposes to perform the separation recursively, separate each new histogram further based on their respective means. Its recursive nature implies scalable preservation which is very useful in consumer electronic products [8].

## E. Minimum Mean Brightness Error Bi-Histogram Equalization in Contrast Enhancement (MMBEBHE):-

This method proposes the novel extension of BBHE referred to as Minimum mean brightness error bi-histogram equalization. It performs the separation based on the threshold level which would return minimum absolute mean brightness error (AMBE) [9]. MMBEBHE works in these three steps defined by the following-

1). First it calculates the AMBE for each of the threshold level.
2). Then it finds the minimum threshold level, $X_T$ that yield minimum MBE,
3). Finally it separates the input histogram based on the $X_T$ found in step 2 and applies histogram equalization each of the separated histograms as in BBHE.

## F. Recursive Sub-Image Histogram Equalization Applied To Gray Scale Image (RSIHE):-

The recursive sub-image histogram equalization (RSIHE) is a contrast enhancement method based on the histogram equalization. This method is generalization of DSIHE. The RSIHE method uses cumulative probability density equal to 0.5 for separating the input histogram into the sub-histogram [10]. This process can be applied recursively for a specified number of recursion levels to divide the image equally into sub-images. Each sub image $X_i$ will contain the same number of pixels. The recursive nature of this method offers scalable brightness preservation.

## G. Recursively Separated and Weighted Histogram Equalization for Brightness Preservation and Contrast Enhancement (RSWHE):-

This is a new histogram equalization method named recursively separated and weighted histogram equalization for brightness preservation and contrast enhancement (RSWHE). To enhance the image contrast as well as preserve the image brightness. RSWHE consists of three modules *Histogram segmentation, Histogram weighting, and Histogram equalization module* [12].

*1. **Histogram segmentation module:-***

This module decomposes the input image histogram in the same way as RMSHE of RSIHE does. It divides the input histogram H(X) recursively up to the some precise recursion level. Histogram segmentation is done on the two ways first one is based on the mean and another based on the median of the sub-histograms.

*2. **Histogram weighting module:-***

Now we know that at recursion level r, the histogram segmentation module has generated $2^r$ sub histograms, i.e. $H_i^r(X), \{1 \leq i \leq 2^r\}$. The histogram weighting module specially modifies the probability density function of each sub histogram using the normalized power law function.

For each sub-histogram $H_i^r(X)$, corresponding original PDF $p(X_k)$, the weighted PDF $p_w(X_k)$, probability density equation is described in (26)-

$$p_w(X_k) = p_{max}\left(\frac{p(X_k)-p_{min}}{p_{max}-p_{min}}\right)^{\alpha_i} + \beta, (L_i \leq k \leq U_i) \qquad (26)$$

Where,

$p_{max}$ and $p_{min}$ are maximum and minimum probability value from original histogram respectively. $\alpha_i$ is an accumulative probability value for i[th] sub-histogram $H_i^r(X)$. $\alpha_i$ is calculated for each sub-histogram.

$$\alpha_i = \sum_{k=L_i}^{U_i} p(X_k) \qquad (27)$$

$\beta$ is a value which is $\geq$ (grater then) 0. The degree of mean brightness and contrast enhancement of output image can be controlled by adjusting $\beta$. It is experimentally found that the output image with the satisfactory quality can be produced, when $\beta$ is around $p_{max}.|X_M - X_G|/(X_{max} - X_{min})$, where $X_{max}$ and $X_{min}$ are the greatest and the least gray levels of the input image$X$, respectively [12]. After weighting process normalization is required to normalize the weighted histogram.

*3. **Histogram equalization Module:-***

The resultant weighted and normalized PDF called $p_{wn}(X_k)$, where the i[th] sub histogram is $H_i^r(X)$ where $(1 \leq i \leq 2^r)$ over the range $[X_{L_i}, X_{U_i}]$. The task of histogram equalization module is to separately equalize each of all $2^r$ sub histogram by using the histogram equalization method. Now finally combining all the resultant sub-images produce output of RSWHE method.

## 2. EXPERIMENTAL RESULTS

We have processed many images to show the performance of different histogram equalization techniques discussed above. We have selected some standard images and apply HE, BBHE, DSIHE, RMSHE, RSIHE, MMBEBHE, RSWHE methods to evaluate their performance. The performance is assessed on the basis of absolute mean brightness error (AMBE), peak signal to noise ratio (PSNR), and structure similarity index measure (SSIM).

- **_Absolute Mean Brightness Error:-_**
  Absolute mean brightness error (AMBE) is used to evaluate Brightness preservation in processed image.

  AMBE is defined as-

  $$AMBE(X,Y) = |X_M - Y_M| \tag{28}$$

  Where, $X_M$ is mean of the input image $X = \{x(i,j)\}$
  And $Y_M$ is mean of the output image $Y = \{y(i,j)\}$.

  Minimum brightness error means batter brightness preservation.

- **_Peak signal to noise ratio:-_**
  For peak signal to noise ratio (PSNR) assume an input image is $X(i,j)$ which contains $M \times N$ pixels and the processed image $Y(i,j)$.

  First compute the Mean Squared Error (MSE),

  $$MSE = \frac{\sum_{i=1}^{M} \sum_{j=1}^{N} |X(i,j) - Y(i,j)|^2}{M \times N} \tag{29}$$

  Now peak signal to noise ratio (PSNR)

  $$PSNR = 10 \log_{10} \frac{(L-1)^2}{MSE} \tag{30}$$

- **_Structured Similarity index:-_**
  The structural similarity index is a method for measuring the similarity between two images. The SSIM index is a full reference metric, in other words, the measuring of image quality based on an initial uncompressed or distortion-free image as reference. Here we define the structural similarity index:

  $$SSI(x,y) = \frac{(2\mu_x\mu_y + C_1)(2\sigma_{xy} + C_2)}{(\mu_x^2 + \mu_y^2 + C_1)(\sigma_x^2 + \sigma_y^2 + C_2)} \tag{31}$$

  Where $\mu_x$ is the mean of image X, $\mu_y$ is mean of Y, $\sigma_x$ is standard deviation of image x, $\sigma_y$ is standard deviation of image y, $\sigma_{xy}$ is squre root of covariance of image x and y, and $C_1$, $C_2$ are contents.

- ***Entropy:-***
  The entropy is a valuable tool to measure the richness of the details in the output image. For a given PDF $p$, entropy Entropy[$p$] is computed by (32)

$$Entropy[p] = -\sum_{k=0}^{L-1} p(X_k) \log_2 p(X_k) \tag{32}$$

***Analysis Results Tables:***

*Table-1: For Mean Brightness Error (AMBE)*

| Images | HE | BBHE | DSIHE | RMSHE (r=2) | RSIHE (r=2) | MMBEBHE | RSHWE-M (r=2) | RSWHE-D (r=2) |
|---|---|---|---|---|---|---|---|---|
| couple | 96.88 | 32.07 | 40.97 | 10.44 | 19.74 | 17.48 | 2.19 | 3.73 |
| einstein | 20.67 | 17.23 | 9.98 | 9.43 | 8.80 | 1.35 | 1.80 | 0.64 |
| f_16 | 52.29 | 0.194 | 18.10 | 2.22 | 4.90 | 0.46 | 2.37 | 0.65 |
| fighterplane | 98.81 | 14.75 | 33.14 | 4.43 | 19.26 | 7.47 | 1.17 | 3.12 |
| fruits | 17.99 | 10.03 | 11.24 | 5.81 | 5.93 | 2.03 | 3.11 | 2.34 |
| girl | 6.70 | 22.60 | 12.89 | 0.65 | 1.04 | 6.10 | 0.22 | 2.58 |
| house | 10.13 | 3.300 | 14.28 | 4.35 | 3.50 | 3.04 | 1.67 | 2.64 |
| lady | 70.98 | 21.20 | 29.21 | 9.14 | 13.80 | 13.74 | 2.51 | 4.05 |
| plane | 37.09 | 1.463 | 23.75 | 9.83 | 5.86 | 7.04 | 2.33 | 1.19 |
| tank | 21.76 | 18.90 | 9.64 | 10.4 | 7.53 | 2.37 | 6.46 | 5.13 |
| *Average* | *43.33* | *14.17* | *20.32* | *6.68* | *9.04* | *6.11* | *2.38* | *2.61* |

*Table-2: For Peak Signal to Noise Ratio (PSNR)*

| Images | HE | BBHE | DSIHE | RMSHE (r=2) | RSIHE (r=2) | MMBEBHE | RSHWE-M (r=2) | RSWHE-D (r=2) |
|---|---|---|---|---|---|---|---|---|
| couple | 7.56 | 13.43 | 12.14 | 19.82 | 15.76 | 19.57 | 35.41 | 30.80 |
| einstein | 15.05 | 15.22 | 16.06 | 19.62 | 19.66 | 18.69 | 27.27 | 26.86 |
| f_16 | 11.49 | 20.16 | 15.63 | 23.37 | 22.01 | 21.87 | 31.93 | 32.71 |
| fighterplane | 6.99 | 15.63 | 12.51 | 22.21 | 15.15 | 20.04 | 32.25 | 28.81 |
| fruits | 16.81 | 18.98 | 18.59 | 24.59 | 22.57 | 23.49 | 32.05 | 31.86 |
| girl | 13.31 | 13.60 | 14.41 | 27.37 | 19.79 | 14.57 | 35.13 | 30.34 |
| house | 17.70 | 17.83 | 17.46 | 21.03 | 20.79 | 19.26 | 29.46 | 30.02 |
| lady | 10.16 | 16.20 | 14.68 | 21.21 | 18.83 | 21.74 | 35.17 | 31.04 |
| plane | 10.97 | 13.32 | 11.66 | 20.59 | 14.97 | 17.23 | 35.03 | 37.69 |
| tank | 13.10 | 13.19 | 14.10 | 17.46 | 17.21 | 15.81 | 24.38 | 24.18 |
| Average | 12.31 | 15.76 | 14.72 | 21.73 | 18.77 | 19.23 | 31.81 | 30.43 |

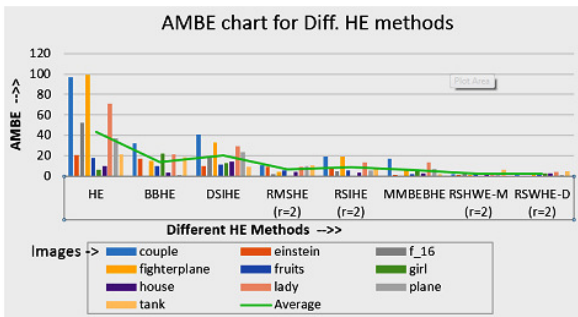

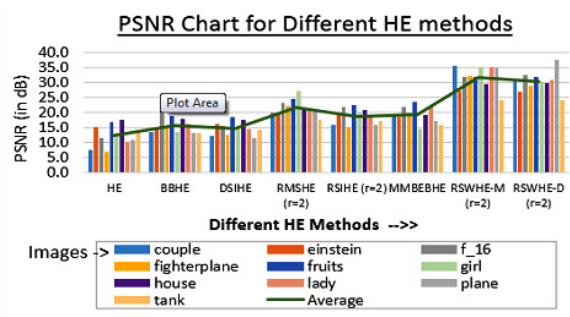

(a)                                         (b)

Fig.3. (a) Shows chart of AMBE for different HE methods and (b) shows chart of PSNR for different HE methods

*Table-3: For Structure Similarity Index (SSI)*

| Images | HE | BBHE | DSIHE | RMSHE (r=2) | RSIHE (r=2) | MMBEBHE | RSWHE-M (r=2) | RSWHE-D (r=2) |
|---|---|---|---|---|---|---|---|---|
| couple | 0.29 | 0.68 | 0.61 | 0.90 | 0.80 | 0.79 | 0.99 | 0.98 |
| einstein | 0.68 | 0.67 | 0.68 | 0.82 | 0.83 | 0.74 | 0.95 | 0.94 |
| f_16 | 0.56 | 0.90 | 0.74 | 0.91 | 0.91 | 0.90 | 0.98 | 0.99 |
| fighterplane | 0.16 | 0.64 | 0.37 | 0.86 | 0.63 | 0.83 | 0.98 | 0.97 |
| fruits | 0.92 | 0.92 | 0.92 | 0.94 | 0.93 | 0.94 | 0.99 | 0.99 |
| girl | 0.30 | 0.38 | 0.35 | 0.94 | 0.65 | 0.37 | 0.99 | 0.97 |
| house | 0.57 | 0.62 | 0.59 | 0.77 | 0.70 | 0.64 | 0.97 | 0.97 |
| lady | 0.61 | 0.87 | 0.80 | 0.94 | 0.91 | 0.93 | 1.00 | 0.99 |
| plane | 0.34 | 0.48 | 0.37 | 0.79 | 0.54 | 0.60 | 0.98 | 0.99 |
| tank | 0.50 | 0.49 | 0.51 | 0.69 | 0.66 | 0.56 | 0.89 | 0.88 |
| *Average* | *0.49* | *0.67* | *0.59* | *0.86* | *0.76* | *0.73* | *0.97* | *0.97* |

*Table-4: For Entropy*

| Images | Original | HE | BBHE | DSIHE | RMSHE (R=2) | RSIHE (R=2) | MMBEBHE | RSWHE-M (R=2) | RSWHE-D (R=2) |
|---|---|---|---|---|---|---|---|---|---|
| Couple | 6.42 | 6.25 | 6.19 | 6.25 | 6.22 | 6.24 | 6.19 | 6.35 | 6.31 |
| Einstein | 6.89 | 6.75 | 6.75 | 6.74 | 6.71 | 6.71 | 6.73 | 6.82 | 6.86 |
| F_16 | 6.70 | 6.44 | 6.60 | 6.53 | 6.56 | 6.53 | 6.61 | 6.68 | 6.63 |
| Fighter-plane | 5.64 | 5.41 | 5.54 | 5.50 | 5.46 | 5.41 | 5.48 | 5.58 | 5.49 |
| Fruits | 7.59 | 7.45 | 7.43 | 7.45 | 7.44 | 7.43 | 7.41 | 7.55 | 7.55 |
| Girl | 5.59 | 5.28 | 5.28 | 5.26 | 5.40 | 5.13 | 5.24 | 5.51 | 5.35 |
| House | 6.50 | 6.26 | 6.25 | 6.22 | 6.22 | 6.24 | 6.23 | 6.45 | 6.47 |
| Lady | 7.05 | 6.90 | 6.90 | 6.91 | 6.89 | 6.83 | 6.82 | 7.01 | 7.00 |
| Plane | 4.00 | 3.88 | 3.93 | 3.89 | 3.97 | 3.95 | 3.92 | 4.00 | 4.00 |
| Tank | 5.99 | 5.88 | 5.87 | 5.87 | 5.94 | 5.93 | 5.85 | 5.99 | 5.98 |
| *Average* | *6.24* | *6.05* | *6.08* | *6.06* | *6.08* | *6.04* | *6.05* | *6.19* | *6.16* |

***Assessment of Brightness Preservation:***

The results shown in the Table-1 presents the performance of brightness preservation of various methods discussed in this paper. Based on the observation of Table-1 we see that RSWHE-M (Mean based) is best in brightness preservation. RSWHE-D (Median based) is second best method for brightness preservation.

***Assessment of contrast enhancement:***

The results shown in the Table-2 presents the PSNR (Peak Signal to Noise Ratio) values for various methods applied to some standard images. PSNR is a metrics for image quality assessment. The greater the PSNR, the better the image quality is. Based on the observation of Table-2 we see that the RSWHE-M (Mean based) has the greater PSNR values of each image and an average PSNR for all images is also has greater value as compared to other histogram equalization methods. The RSWHE-D also produces better results as compared to other methods.

Table 4, presents the Entropy values compared with original image entropy to other HE based methods for some standard images. Entropy is used to measure the richness of details in image. RSWHE-M and RSWHE-D performs batter then others in terms of entropy.

### Inspection of visual quality:

In addition with brightness preservation and contrast enhancement an Image quality is also an important factor in image processing. The processed image should be visually acceptable to human eye and should have natural appearance.

We have tested no. of images with all of methods discussed in this paper. Some of them are presented here. Figure-4 shows original image of "Couple" with processed images by HE, BBHE, DSIHE, RMSHE, RSIHE, MMBEBHE, RSWHE-M (Mean based) and RSWHE-D (Median based) methods. Figure-5 shows the result images of "Einstein".

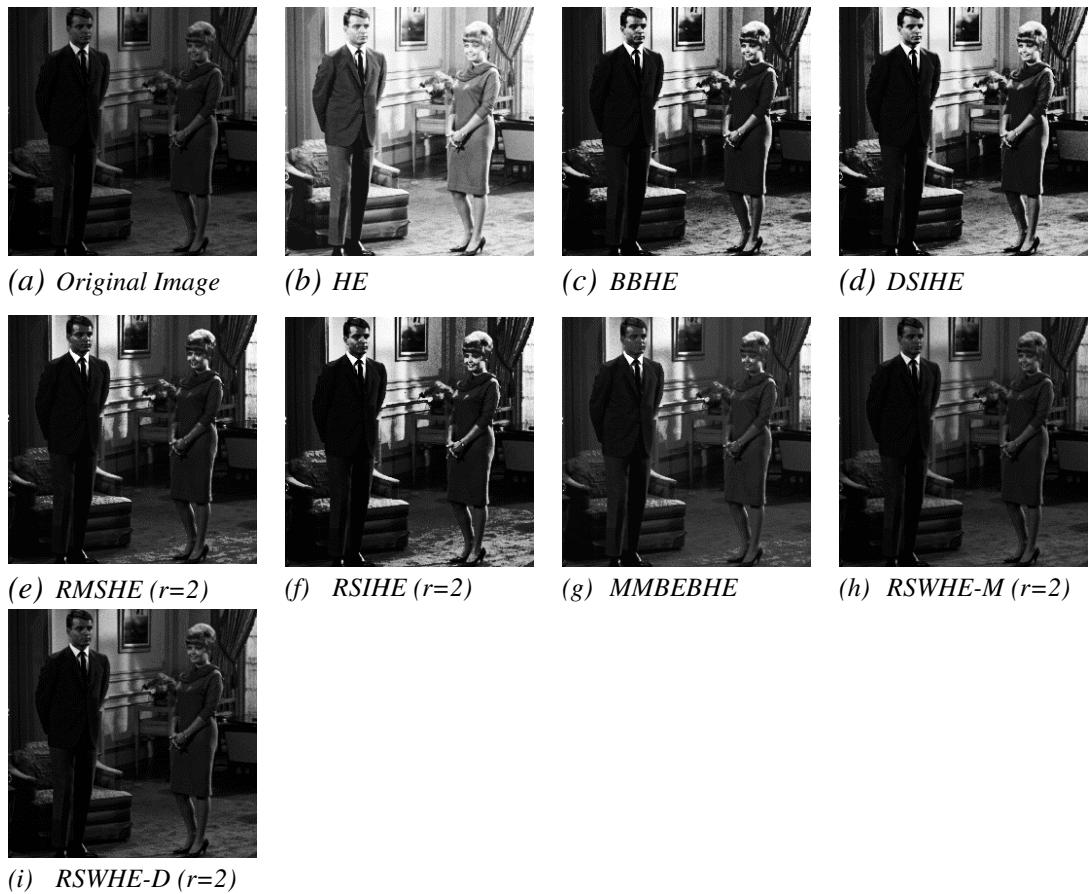

(a) Original Image     (b) HE     (c) BBHE     (d) DSIHE

(e) RMSHE (r=2)     (f) RSIHE (r=2)     (g) MMBEBHE     (h) RSWHE-M (r=2)

(i) RSWHE-D (r=2)

Fig. 4. Results of all methods tested on the image of 'Couple'.

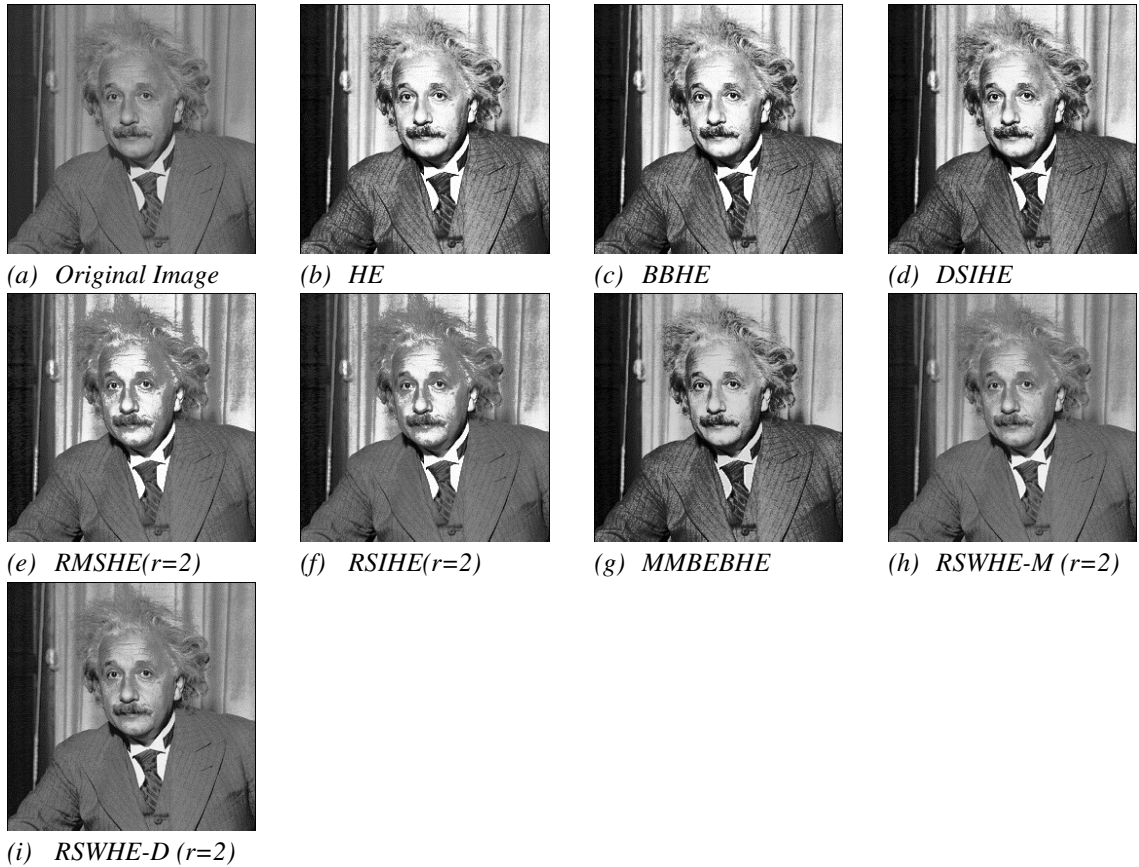

Fig. 5. Results of all methods tested on the image of 'Einstein'.

## 3. CONCLUSION

This paper presents comparative study of different histogram equalization based image enhancement methods. After observation of experimental results for brightness preservation we observed that the brightness preservation is not handled well by HE, BBHE and DSIHE, but it can be handled properly by RMSHE, RSIHE, and RSWHE. RSWHE-M offers better brightness preservation, better contrast enhancement and better structure similarity index as compared to other methods. RSWHE-D is the second best method for brightness preservation and contrast enhancement. The RSWHE method also offers scalable brightness preservation because of its recursive nature. From our study it is observed that a huge work has already been done in this field but still there exist much space for future work.

**ACKNOWLEDGMENT**

We would like to thank Dr. Song Der Chen, Putra Malaysia Univ., Serdang, Malaysia for giving help in MMBEBHE[9].

## AUTHORS

**Omprakash Patel**

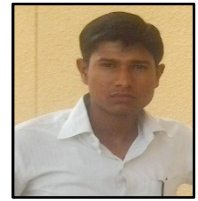

He was born in Jabalpur, India. He received his Bachelor of Engineering degree in Computer Science & Engineering from GGITS Jabalpur affiliated from RGPV Bhopal, in 2010. He is currently pursuing Master of Technology (M.Tech.) in Computer Science & Application from School of Information Technology, UTD, RGPV, Bhopal, India. His research interest includes image enhancement and medical image processing.

**Yogendra P.S. Maravi**

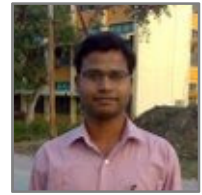

He obtained his Bachelor's degree in Information Technology from UIT, RGPV, Bhopal. He received his M.Tech degree in Computer Science from School of Computer Science, DAVV, Indore. He is working as Assistant Professor in School of Information Technology, UTD, RGPV, Bhopal, India.

**Dr. Sanjeev Sharma**

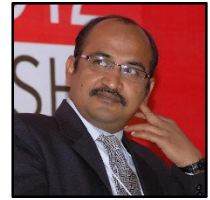

He is working as Associate Professor in the School of Information Technology, UTD, RGPV, Bhopal, India. His research interest is in Mobile Computing. He is a member of Computer Society of India ACM, Computer Science Teachers Association (CSTA).